%% file: esapub2005.tex
          [2001/04/25 1.1 (PWD)]
\documentclass[a4paper,twocolumn]{esapub2005} 
\usepackage[utf8]{inputenc}
\usepackage{times}
\usepackage[square,numbers]{natbib}
\usepackage{url}
\usepackage{xcolor}
\usepackage{amsmath}
\usepackage{amssymb}
\usepackage{amsthm}
\usepackage{booktabs}
\usepackage{layouts}
\usepackage{siunitx}
\usepackage{multirow}
\usepackage{graphicx}
\usepackage{placeins}
\usepackage{optidef}
\usepackage{stmaryrd}
\title{An adaptive hierarchical control framework for quadrupedal robots in planetary exploration}
\author[1,$\star$]{Franek Stark}
\author[1,$\star$]{Rohit Kumar}
\author[1]{Shubham Vyas}
\author[2]{Hannah Isermann}
\affil[1]{Robotics Innovation Center, DFKI GmbH, Bremen, Germany, \small{\{franek.stark, r.kumar, shubham.vyas, dennis.mronga, frank.kirchner\}@dfki.de}}
\author[3]{Jonas Haack}
\author[3]{Mihaela Popescu}
\author[1,3]{Jakob Middelberg}
\affil[2]{German Aerospace Center, Institute of Space Systems, Bremen, Germany, hannah.isermann@dlr.de}
\affil[3]{University of Bremen, Germany, \small{\{jhaack, mpopescu, middelberg\}@uni-bremen.de}}
\author[1]{Dennis Mronga}
\author[1,3]{Frank Kirchner}
\affil[$\star$]{Franek Stark and Rohit Kumar contributed equally to this work}

\newcommand{\ra}[1]{\renewcommand{\arraystretch}{#1}}
\input{acros}
\usepackage{hyperref}

\begin{document}


\maketitle

\begin{abstract}
Planetary exploration missions require robots capable of navigating extreme and unknown environments. While wheeled rovers have dominated past missions, their mobility is limited to traversable surfaces. Legged robots, especially quadrupeds, can overcome these limitations by handling uneven, obstacle-rich, and deformable terrains. However, deploying such robots in unknown conditions is challenging due to the need for environment-specific control, which is infeasible when terrain and robot parameters are uncertain. This work presents a modular control framework that combines model-based dynamic control with online model adaptation and adaptive footstep planning to address uncertainties in both robot and terrain properties. The framework includes state estimation for quadrupeds with and without contact sensing, supports runtime reconfiguration, and is integrated into ROS 2 with open-source availability. Its performance was validated on two quadruped platforms, multiple hardware architectures, and in a volcano field test, where the robot walked over 700~m.
\end{abstract}

\section{Introduction}
\input{sections/intro}

\section{Adaptive walking controller}
To enable flexible deployment of the adaptive quadruped controller in different environments and hardware configurations, we implemented the control framework into ROS~2. 
\begin{figure}[htbp]
    \centering
    \includegraphics[width=\linewidth]{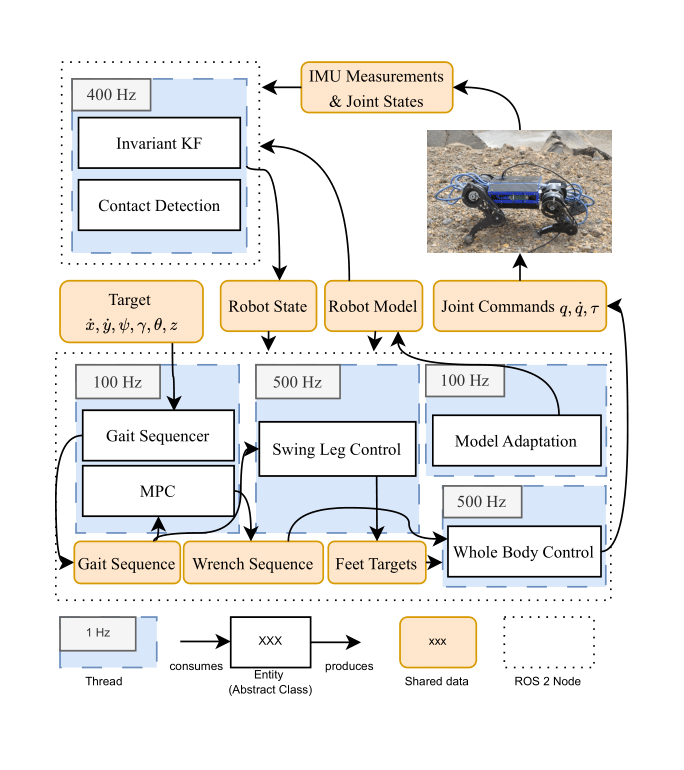}
    \caption{Block diagram showing the architecture of the control framework}
    \vspace{-0.3cm}
    \label{fig:control_framework}
\end{figure}
\autoref{fig:control_framework} shows the sub components.
It is mainly split into two ROS~2 nodes. The controller node and the state estimation, which both are explained in the following.

The dynamic walking controller node follows the hierarchical model based control approach presented by \cite{di_carlo_dynamic_2018, kim_highly_2019}, which originally consisting of four components:
(1) The \textit{\gls{gs}}, based on the control target, determines a gait sequence, containing the foot contact plan along with the robot's base target trajectory over the prediction horizon $N$.
(2) The optimal contact forces that are needed to track the planned robot body trajectory are calculated using \textit{\gls{mpc}} on \gls{srbd}. 
(3) The \textit{\gls{slc}} computes the trajectory from one footstep location to the next one, using Bézier curves.
(4) The \textit{\gls{wbc}} solves for the joint accelerations and torques that satisfy the desired foot trajectories, body posture and velocity, and contact forces, while ensuring consistency with the full-body dynamics.
In our controller, a (5) model adaptation component is added, which estimates the model parameters online and then updates them in the other components.
The controller node hosts the controller sub components which are distributed over several threads, allowing the parallel executions of the control loops over the different hierarchies as indicated in \autoref{fig:control_framework}.
Each sub components interface is defined as an \textit{pure virtual class} from which different implementations can be derived. Different implementation of a sub component can than therefore be switched during runtime.
For the GS we use two different implementation a base-lime \gls{sgs} and an advanced \gls{ags} which is explained in \autoref{subsec:ags}.
The (2) \gls{mpc} is formulated as described in \cite{di_carlo_dynamic_2018, kim_highly_2019} and so is the (3) SLC. The (4) \gls{wbc} uses a slightly different approach than presented in \cite{kim_highly_2019} which is explained in \autoref{subsec:wbc}. 
The (5) model adaptation is explained in \autoref{subsec:ma}.

For the state estimation we adopt a contact-aided invariant EKF (InEKF)~\cite{inekf2020} for the floating-base state and a momentum-based contact detector for detecting contacts~\cite{bledt2018contact} as explained in more details in \autoref{subsec:state}. Together, these methods offer a good combination for estimating the state of locomotion on rough, uncertain terrain. 

\subsection{Adaptive Gait Sequencer}\label{subsec:ags}
In order to stabilize the gait and make it cost efficient, an \gls{ags} was implemented. The \gls{ags} works in two stages. First, the gait parameters (duty factor $d$, phase offset $\theta_i$ and period $T$) are selected based on the current velocity $v$ of the robot. In the second stage, the current phase $\phi_i$ for each leg $i$ gets updated and corrected. 
In the first stage, a bio-inspired stride length \cite{holmesDynamicsLeggedLocomotion2006} was used to calculate the period:
\begin{align}
    T =& \frac{l_\text{stride}}{v}   \\
    l_\text{stride} =& 2.3 (Fr)^{0.3} \cdot h,
\end{align}
where $h$ is the height of the COM and $Fr = v^2/g\,h$ is the Froude number with $g = 9.81$\,m/s\textsuperscript{2}.
To determine the duty factor, we followed the findings of \cite{maesSteadyLocomotionDogs2008a}, who reported that swing times in Belgian shepherd dogs (\textit{Canis familiaris L.}) remain approximately constant across velocities and gaits. Accordingly, the duty factor was computed by enforcing a constant swing time of $t_\text{swing}=0.2$\,s:
\begin{equation}
    d = \frac{T - t_\text{swing}}{T}.
\end{equation}
The phase offsets are selected for trotting gait as $\theta_\text{FL} = 0$, $\theta_\text{FR} = 0.5$, $\theta_\text{HL} = 0.5$ and $\theta_\text{HR} = 0$ for the four legs (F: front, H: hind, L: left, R: right).

In the second stage, the phase gets first updated. However, 
as the continuous gait parameter adaptation would lead to inconsistent contact states, the phase update has to ensure their consistency:
\begin{equation}
    \phi_{i,t} = \begin{cases}
        \Tilde{\phi_{i,t}} \cdot \dfrac{d_{i,t}}{d_{i,t-1}} & \hspace{-5mm}\text{if~} \Tilde{\phi_{i,t}} < d_{i,t-1}\\[4mm]
        \dfrac{(\Tilde{\phi_{i,t}} - d_{i,t-1}) \cdot (1-d_{i,t})}{1-d_{i,t-1}} + d_{i,t} & ~~\text{otherwise}
    \end{cases}
\end{equation}
with
\begin{equation}
    \label{eq:first_phase_update}
    \Tilde{\phi_{i,t}} = \phi_{i,t-1} + \frac{\Delta t}{T}
\end{equation}
where $t$ symbolizes the time and $\Delta t$ the duration of one time step. 
This phase update leads to a drift in the phase offsets between the legs, therefore, the phase offsets get corrected in a second step. 
For this, we have to define a reference phase $\phi_\text{ref}$. Here, the leg with the longest remaining swing time was chosen, as the phase during swing phase cannot be adapted to ensure the constant swing time. If all legs are in contact, the leg which is closest to the swing phase is used. 
With this reference phase, we can calculate the current phase offset $\Tilde{\theta_i}$ for each leg:
\begin{equation}
    \Tilde{\theta_i} = \phi_\text{ref} + \theta_\text{ref} - \phi_i
\end{equation}
and the phase offset error $e^\theta_i \in [-0.5,\,0.5]$:
\begin{equation}
    e^\theta_i = ((\Tilde{\theta_i} - \theta_i + 0.5) \mod{1.0}) - 0.5.
\end{equation}
To interpolate the phases smoothly, a maximum error correction step $\hat{e}$ was defined as
\begin{equation}
    \hat{e} = \frac{\Delta t}{2 T},
\end{equation}
ensuring a full correction within two gait cycles. 
Finally, the phase of all legs in stance phase is updated using:
\begin{equation}\label{eq:phase_offset_correction}
    \phi_i = \begin{cases}
    \min\left(\phi_i + \min(e^\theta_i,\, \hat{e}),\, d_i\right)  & ~~\text{if~}e^\theta_i \geq 0\\
    \max\left(\phi_i + \max(e^\theta_i,\, -\hat{e}),\, 0\right)  & ~~\text{otherwise,}
    \end{cases}
\end{equation}
avoiding transitions between the contact states.
As this phase offset correction approach ensures smooth interpolation between different phase offsets, it can also implicitly transition between different gait types, which are commanded as the new desired phase offset $\theta_i$.\\
With the duty factor and the phase, the contact state of each leg can be calculated.
Leg $i$ is in contact when:
\begin{equation}
    \phi_i < d.
\end{equation}

As reference, a simple gait sequencer (SGS) with constant gait parameters ($d = 0.6$, $T= 0.5$\,s and the same $\theta_i$ as the \gls{ags} was used.

\subsection{Whole Body Control (WBC)}\label{subsec:wbc}
For stabilizing the feet and COM trajectories we formulate the following the QP:
{
\footnotesize
\begin{subequations}
\begin{eqnarray}
\label{eq:acceleration_wbc}
 \def\arraystretch{1.5}
\underset{\ddot{\mathbf{q}}, \boldsymbol{\tau}, \mathbf{f}}{\text{min}} & \left\|\sum_i  \mathbf{w}_i^T\left( \mathbf{J}_i\ddot{\mathbf{q}} + \dot{\mathbf{J}}_i\dot{\mathbf{q}} - \dot{\mathbf{v}}^d_i\right) + \sum_j\mathbf{w}_j^T\left(\mathbf{f}_j^d - \mathbf{f}_j\right) \right\| \label{eq:acceleration_wbc_cost}\\
\text{s.t.} & \mathbf{H}\ddot{\mathbf{q}} + \mathbf{h} = \mathbf{S}^T\boldsymbol{\tau} + \sum_j\mathbf{J}_j^c\mathbf{f}_j \label{eq:acceleration_c1}\\
      & \mathbf{J}_j^c\ddot{\mathbf{q}} = -\dot{\mathbf{J}}_j^c\dot{\mathbf{q}}, \quad \forall j\label{eq:acceleration_c2} \\
       & \boldsymbol{\tau}_{m} \leq \boldsymbol{\tau} \leq \boldsymbol{\tau}_{M} \label{eq:acceleration_c3}\\
       & \mu f_z\leq |f_x|, \,\, \mu f_z\leq |f_y|, \,\, f_z>0 \label{eq:acceleration_c4}     
\end{eqnarray}
\end{subequations}
}Here, $\dot{\mathbf{q}},\ddot{\mathbf{q}} \in \mathbb{R}^{6+n}$ are the robot joint velocities and accelerations, $\boldsymbol{\tau} \in \mathbb{R}^{n}$ the joint torques, $n$ the number of actuated robot joints, $\mathbf{J}_i,\dot{\mathbf{J}}_i \in \mathbb{R}^{6 \times n}$ the Jacobian and its derivative for the $i$-th task, $\mathbf{w}_i \in \mathbb{R}^m$ the respective task weights, $m$ the number of task variables, $\dot{\mathbf{v}}_i^d \in \mathbb{R}^m$ the desired task space acceleration for the $i$-th task, $\mathbf{H} \in \mathbb{R}^{n \times n}$ the joint space mass-inertia matrix, $\mathbf{h} \in \mathbb{R}^{n}$ the vector of Coriolis-centrifugal forces, $\mathbf{S} \in \mathbb{R}^{(6+n) \times n}$ the actuator selection matrix, $\mathbf{J}^c \in \mathbb{R}^{6 \times n}$ the contact Jacobian, $\boldsymbol{\tau}_M,\boldsymbol{\tau}_m \in \mathbb{R}^{n}$ the upper and lower joint torque limits, $\mu \in \mathbb{R}$ the contact friction coefficient and $\mathbf{f}^d,\mathbf{f} \in \mathbb{R}^{3}$ the desired and actual feet contact forces.
The cost function is designed to minimize the error between desired and actual task space accelerations for the given feet and COM trajectories, as well as the error between desired and actual contact forces. 
The constraints of the QP ensure compliance with the equations of motion (\ref{eq:acceleration_c1}), non-moving ground contacts (\ref{eq:acceleration_c2}),  torque limits (\ref{eq:acceleration_c3}), and frictional constraints (\ref{eq:acceleration_c4}). We use a polyhedral friction cone approximation to model the static ground friction. 
Trajectory stabilization is performed in task space, where the desired task space accelerations $\ddot{\mathbf{v}}_i^d$ for the $i$-th task (which can be either COM or feet trajectory stabilization) are computed using PD-control:
\begin{equation}
\dot{\mathbf{v}}_i^d = \dot{\mathbf{v}}_i^r + \mathbf{k}^T_{d,i}(\mathbf{v}^r_i-\mathbf{v}_i) + \mathbf{k}_{p,i}^T(\mathbf{x}^r_i\ominus\mathbf{x}_i)   
\end{equation}
Here, $\dot{\mathbf{v}}^r,\mathbf{v}^r \in se(3)$ and $\mathbf{x}^r \in SE(3)$ are the reference spatial 
acceleration, twist and pose for the COM and for the feet, and $\mathbf{k}_{d}, \mathbf{k}_{p}$ are the proportional and derivative gains. 
Since we also want to control the floating base orientation, the pose control error must be computed using the matrix logarithm, which is denoted by the $\ominus$ symbol.
The reference spatial acceleration, twist and pose for the COM and feet, as well as the reference contact forces are computed by the MPC and SLC resepctively. 

\subsection{Kalman Filtering (KF) for Model Adaptations}\label{subsec:ma}
For identifying the mass and center of mass (COM) of the robot, when a payload is added, we use a KF. It utilizes the floating base dynamics, state information, and ground reaction forces (GRF) to estimate the mass and COM x- and y-components of the entire system. As the COM z-component is unobservable in default configuration, it is assumed to be $0$. The dynamics in simplified form can be written as follows:
\begin{align}
		&m(\dot{\mathbf{v}} + \mathbf{g}) = \sum_{i=0}^{3}{\mathbf{F}_i} \label{eq:EOM1}\\
		&\mathbf{c}\times m\mathbf{g}= \sum_{i=0}^{3}(\mathbf{r}_i\times\mathbf{F}_i),
		\label{eq:EOM2}
\end{align}
where $m$ is the robot's total mass, $\dot{\mathbf{v}}\in \mathbb{R}^3$ the linear acceleration, $\mathbf{g} = \begin{bmatrix}0, 0, g\end{bmatrix}^T\in \mathbb{R}^3$ the gravity vector and $\mathbf{F}_i\in \mathbb{R}^3$ the GRF at foot $i$. $\mathbf{c} \in \mathbb{R}^3$ is the COM position and $\mathbf{r}_i$ the position of foot $i$. All vectors are represented in the inertial frame. Since the robot is generally not expected to perform high rotational acceleration movements, the influence of the second moments of inertia is neglected. The foot positions $\mathbf{r}_i$ and GRFs $\mathbf{F}_i$ are obtained from leg kinematics and inverse dynamics:
\begin{align}
     & \mathbf{r}_i = \mathbf{f}_i(\mathbf{q}_i)\\ 
    & \mathbf{F}_i = (\mathbf{J}^T_i)^{-1}\boldsymbol{\tau}_i.\label{eq:inv_dyn}
\end{align}
$\mathbf{f}_i:\mathbb{R}^3\rightarrow\mathbb{R}^3$ denotes the kinematics of leg $i$ dependent on its joint positions $\mathbf{q}_i \in \mathbb{R}^3$. $\mathbf{J}_i \in \mathbb{R}^{3\times3}$ is the Jacobian matrix of leg $i$ and $\boldsymbol{\tau}_i \in \mathbb{R}^3$ its corresponding vector of joint torques, which are obtained from motor current measurements.

The dynamic parameters of this model can be concatenated in the parameter vector:
\begin{equation}
    \boldsymbol{\pi} = \begin{bmatrix}
        m & mc_x & mc_y
    \end{bmatrix}^T \in \mathbb{R}^{3}. \label{eq:param_vec}
\end{equation}
Applied to the problem of estimating $\boldsymbol{\pi}$ the KF equations are as follows:
\begin{align}
    &\hat{\boldsymbol{\pi}}^-_k = \hat{\boldsymbol{\pi}}^+_{k-1}\label{eq:KF1}\\
    &\mathbf{P}^-_k = \mathbf{P}^+_{k-1} + \mathbf{Q}\label{eq:KF2}\\
    &\mathbf{K}_k = \mathbf{P}^-_k\boldsymbol{\Phi}_k^T(\boldsymbol{\Phi}_k\mathbf{P}^-_k\boldsymbol{\Phi}_k^T + \mathbf{R})^{-1}\label{eq:KF3}\\
    &\hat{\boldsymbol{\pi}}_k^+ = \hat{\boldsymbol{\pi}}_k^- + \mathbf{K}_k(\mathbf{z}_k-\boldsymbol{\Phi}_k\mathbf{\hat{\boldsymbol{\pi}}_k^-})\label{eq:KF4}\\
    &\mathbf{P}_k^+ = (\mathbf{I} - \mathbf{K_k\boldsymbol{\Phi}_k})\mathbf{P}_k^-,\label{eq:KF5}
\end{align}
where $\boldsymbol{P}\in \mathbb{R}^{3\times 3}$ is the estimation covariance matrix, $\boldsymbol{Q}\in \mathbb{R}^{3\times 3}$ the process noise covariance matrix, $\boldsymbol{K}\in \mathbb{R}^{3\times 6}$ the Kalman Gain, $\boldsymbol{\Phi} \in \mathbb{R}^{6\times3}$ the Jacobian of the floating base dynamics with respect to the parameter vector, $\boldsymbol{R}\in \mathbb{R}^{6\times 6}$ the measurement noise covariance matrix, and $\boldsymbol{z}\in \mathbb{R}^6$ is the measurement vector defined by the right-hand sides of~\autoref{eq:EOM1} and ~\autoref{eq:EOM2}. Superscript $-$ and $+$ denote quantities before and after the update step, and subscript $k$ indicates a time step. A more detailed description of the derivations and algorithm can be found in~\cite{haack2025adaptivemodelbasecontrolquadrupeds}.

\subsection{State Estimation with Floating Base and Contacts in Planetary Scenarios}\label{subsec:state}

\input{sections/state_estimation}

\section{Results}
The framework has been open sourced\footnote{https://github.com/dfki-ric-underactuated-lab/dfki-quad} and tested with two different quadrupedal systems (\autoref{fig:quad_vulcano}): a Unitree Go2 (equipped with foot contact sensors) and a custom-built platform (without foot contact sensors). 
Further, the joint command and sensor measurement messages can be rerouted to a simulation, which allows for verifying controller implementations, debugging message handling, and comparing real-world experiments against simulated counterparts under identical interfaces. The control framework was evaluated with respect to computational performance, adaptive gait planning and model adaptation. 
As optimization-based modules such as MPC and WBC rely on efficient QP solving, we first summarize a previously published benchmark of different solver–hardware-QP combinations to assess their real-time suitability. 
We then present results on adaptive gait sequencing from a field test on vulcano, followed by a brief overview of our earlier published work on online model adaptation for mass and center of mass estimation using this framework.
Since all results have been obtained using this control framework, either in simulation or on the real systems, they serve as proof of its practicality and effectiveness under both laboratory and field conditions.

\subsection{Computational evaluation}
Running the described control framework onboard the quadruped requires fast and efficient computational hardware.
The Quadratic Programs (QPs) which have to be solved within the MPC and WBC are the computational most expensive operation in the proposed control framework. 
On limited hardware as it is the case for the onboard computers of quadrupedal robots, the choice of the QP solver is crucial as a slow solver might degrade the control frequency and hence stability of the system.
Another requirement is the energy consumption as not only the computational hardware but also the electric energy might be limited.
There are abundant implementations of QP-Solvers available, which also require different formulations, so choosing the best suitable QP-Solver can be challenging, since studies on the computational efficiency of QP-Solvers on different computer architectures are lacking.
In order to provide a recommendation on which solver, formulation and hardware to choose, in our previous work we introduce the Solve Frequency per Watt (SFPW) performance metric (\ref{eq:sfpw_calculation}), to enable a fair, cross-hardware comparison \cite{stark_benchmarking_2025}:
\begin{gather}
    \frac{{\mathrm{solve~time}}^{-1}}{\mathrm{CPU~power~consumption}} \; \; \; \left[\frac{Hz}{W}\right]
    \label{eq:sfpw_calculation}
\end{gather} %
The performance benchmarks were run on x86 (Desktop and LattePanda) and ARM (Jetson Orin) computers.
We compared different solvers, representing solving techniques like \textit{active set} or \textit{interior point} method.
Table \ref{tbl:efficiency_all} shows the summarized SFPW results for the most efficient solver(s) for different problem sizes and over the target computers from \cite{stark_benchmarking_2025}.
Remarkably, the \gls{ipm} solver \gls{hpipm} \cite{frison_hpipm_2020} shows the highest efficiency for the large QPs on all systems, except for the Full TSID WBC problem solved with the Jetson Orin, on which the \gls{asm} solver Eiqaudprog slightly outperforms.
For the small QPs that arise in the reduced TSID WBC problem, Eiquadprog is the most efficient solver.
\FloatBarrier
\vspace{-1em}
\begin{table}[h]
\centering
\caption{Mean SFPW of MPC problem solvers (\unit{\hertz\per\watt}) for different prediction horizons N}
\vspace{.5em}
\label{tbl:efficiency_all}
\ra{1.3}
\small
\begin{tabular}{@{}cclccc@{}}\toprule
 & & & Jetson & Desktop & LatteP.  \\ 
\cmidrule{1-6}
\parbox[t]{1mm}{\multirow{2}{*}{\rotatebox[origin=c]{90}{MPC}}} & 
\tiny{$N=10$}
&\footnotesize{HPIPM}
& \textbf{193}  & \textbf{61}  &  \textbf{86}  \\
\cmidrule{2-6}
&
\tiny{$N=20$}
&
\footnotesize{HPIPM}
& \textbf{82}  & \textbf{28}  & \textbf{39}  \\
\midrule
\parbox[t]{1mm}{\multirow{3}{*}{\rotatebox[origin=c]{90}{WBC}}} & 
\multirow{2}{*}{\tiny{Full TSID}}
& 
 \footnotesize{Eiquadprog} & \textbf{897}  & 204  & 254 \\
&&\footnotesize{HPIPM} & \textbf{893}  & \textbf{251}  & \textbf{283} \\
\cmidrule{2-6}
&
\tiny{Red. TSID}
& 
\footnotesize{Eiquadprog} & \textbf{1158} & \textbf{305}  & \textbf{334} \\
 \bottomrule
\end{tabular}
\vspace{-1em}
\end{table}
\FloatBarrier
Out of the examined computers, the Jetson Orin (ARM) is almost twice as efficient as the LattePanda (x86) and about three times as efficient as the Desktop PC (x86). All results, comparing solvers and smaller increments between formulations can be found in \cite{stark_benchmarking_2025}.

\subsection{Adaptive Gait Sequencer}
\begin{figure}[htb!]
    \centering
    \includegraphics[width=1\linewidth]{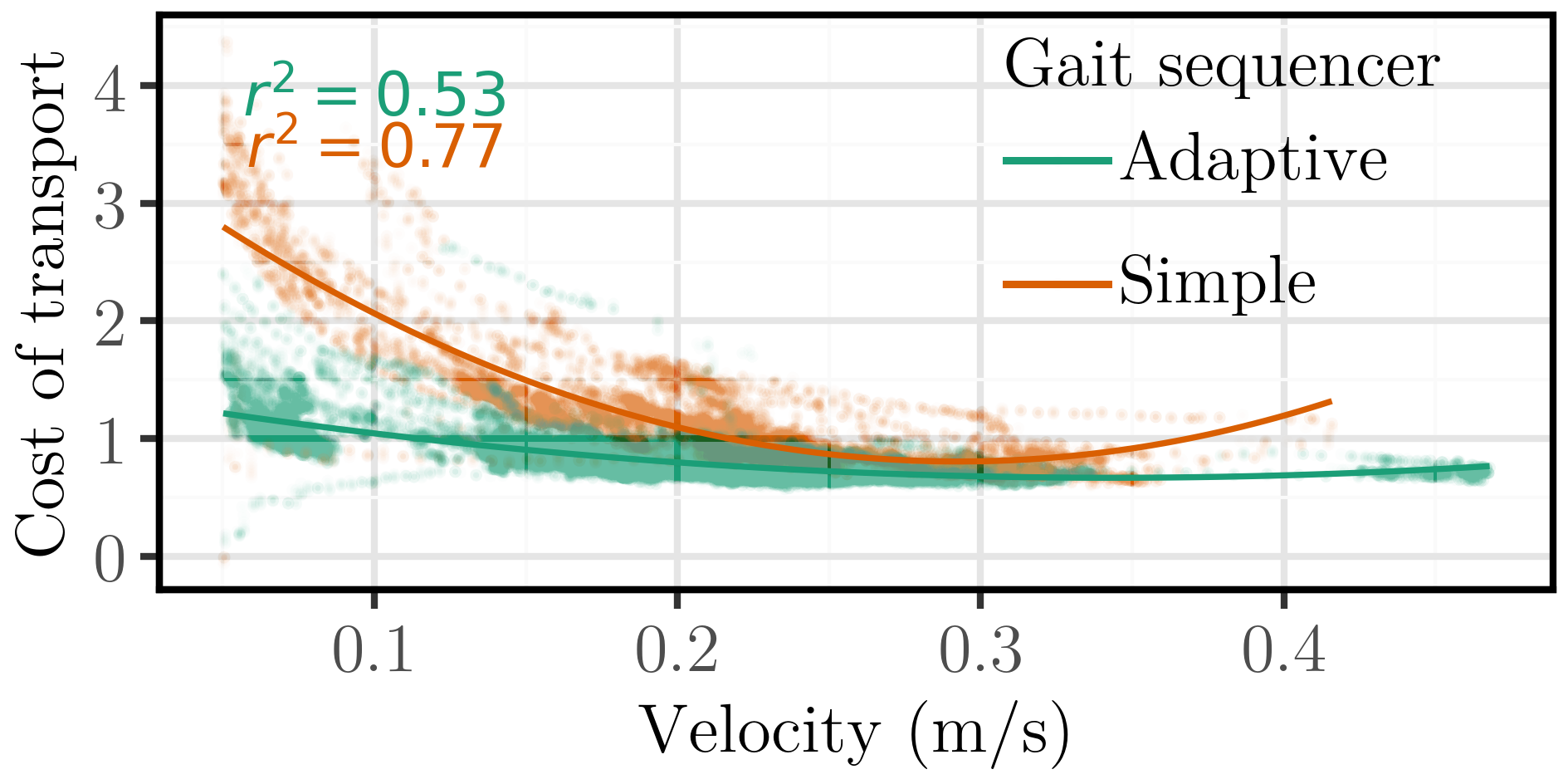}
    \caption{Cost of transport over velocity of the two gait sequencers during trotting on the field trip. The plot shows the combined data of all experiments. The $r^2$ values for each fit are plotted onto the graph.}
    \label{fig:cot_GS}
\end{figure}

During a field test on the crater rim of Vulcano Island, Italy, we conducted experiments to compare the performance of the \gls{ags} and a baseline \gls{sgs}, which works on fixed gait timings, in rough terrain. 
During the experiments, the \gls{ags} only fell twice, resulting on average in a fall every 133.5\,m. In contrast, the \gls{sgs} had more difficulties in the terrain and fell on average every 6.1\,m. By comparing the cost of transport during the experiments (Fig.\,\ref{fig:cot_GS}) we also see an increase in efficiency of the \gls{ags} over the SGS for all velocities.

\subsection{Model Adaptation}
The model adaptation has been evaluated in an experiment where different payloads are placed on the quadrupedal robot over time.
\begin{figure}[!htbp]
    \centering
    \includegraphics[width=\linewidth]{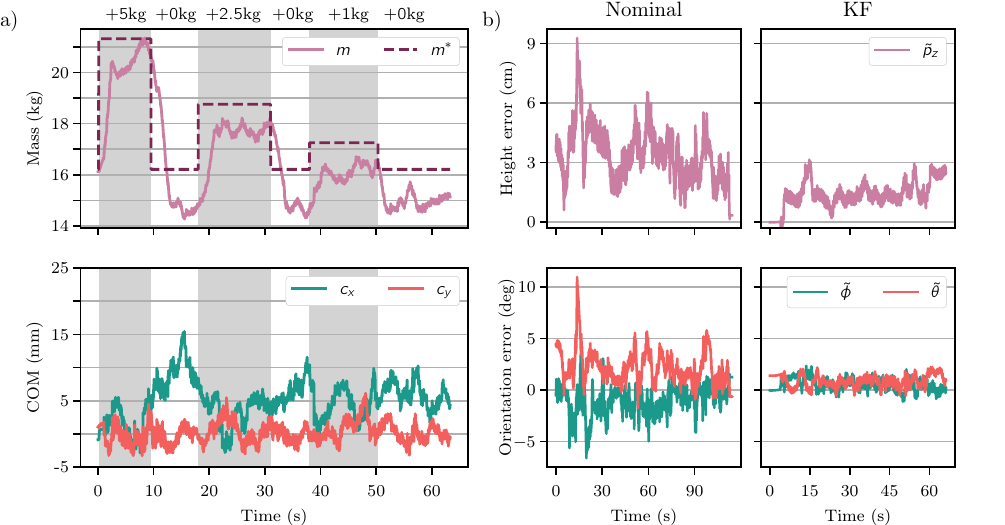}
    \caption{a) Estimated total mass with ground truth (top) and estimated COM (bottom) during dynamic payload switching experiment. The original robot weight is \SI{16.21}{\kg} and the COM offsets are $c_x =$ \SI{8.8}{\mm}, and $c_y =$ \SI{0.0}{\mm}. The time intervals where a specific amount of payload (displayed above the upper graph) is attached to the robot are highlighted in grey. b) Height and orientation error of nominal controller (left) and adaptive controller with KF (right) over dynamic weight switching experiment.}
    \label{fig:weight_switching}
\end{figure}
\autoref{fig:weight_switching}a) shows the estimated mass $m$ the corresponding ground truth $m^\ast$ and estimated COM x and y-components $c_x$ and $c_y$ over the experiment duration. 
No ground truth data is available for the COM offsets. Once converged, the mass estimate shows noise of approximately $\pm$\SI{500}{\g}.
Larger deviations occur when the robot performs abrupt movements to compensate for drift. A bias of approximately \SI{1}{\kg} can be observed. 
Upon adding some payload, the estimate quickly rises by the correct amount. 
When the weight is removed, the estimate converges to a value lower than the original mass. 
The estimate of $c_x$ fluctuates by approximately $\pm$\SI{5}{\mm} while $c_y$ does so at a lower degree around $\pm$\SI{2.5}{\mm}.
The weight changes cannot be observed as clearly in the COM estimation. ~\autoref{fig:weight_switching} b) shows the tracking errors of the nominal and adaptive controller over the same experiment. The adaptation improves the tracking of height and orientation, maintaining a more consistent error across all payloads.
Further results and analysis and a comparison to base-line least squares estimation can be found in \cite{haack2025adaptivemodelbasecontrolquadrupeds}.

\subsection{Discussion and Conclusion}
The adaptive hierarchical control framework presented in this work has shown promising results for quadrupedal locomotion in uncertain environments. By combining adaptive gait sequencing and online model adaptation within a layered architecture, the framework effectively compensates for uncertainties in both robot dynamics and terrain conditions. However, there are many limitations of the current results. First, while the adaptive gait sequencer substantially improved stability, its validation was limited to a planetary analog environment in Vulcano, Italy. Performance in other terrains, such as low-gravity or highly deformable soils, remains to be tested.
Second, the online model adaptation via Kalman filtering, though effective in estimating mass and COM changes, exhibited noise and a systematic bias. These deviations, particularly during abrupt maneuvers, could degrade the accuracy of control over long durations, and the reliable estimation of inertial parameters remains an open challenge in the literature. Finally, the experiments relied on manually switching between different control modules to study the effect of critical subcomponents. While this approach provided valuable insights, future work should focus on making these transitions autonomous by ensuring each subcomponent is sufficiently robust and equipped with fallback strategies.

Despite these limitations, the field trials on Vulcano Island highlight the advantages of embedding adaptability into multiple layers of control. The reduction in falls, improved cost of transport, and increased tracking accuracy under payload changes underline the necessity of combining bio-inspired gait adaptation, MPC, WBC, and online parameter estimation. These results emphasize that resilience in quadrupedal locomotion cannot be achieved by a single control module in isolation, but requires coordinated adaptation across the hierarchy.

In conclusion, this work demonstrates that an adaptive hierarchical control framework can significantly enhance the robustness of quadrupedal robots in planetary exploration contexts. The modular design allows for runtime reconfiguration, integration of diverse sensing strategies, and efficient deployment across different hardware platforms. Future research should address the limitations identified by extending validation to diverse planetary analog environments, refining model adaptation algorithms to reduce bias and noise, and incorporating additional sensing modalities such as vision or tactile feedback. Ultimately, the integration of adaptive hierarchical control with higher-level autonomy and perception modules will be key to enabling fully self-sufficient robotic explorers capable of long-term operation in extraterrestrial environments.

\section*{Acknowledgments}

This work partially funded by the projects: AAPLE (grant number 50WK2275) funded by the German Federal Ministry for Economic Affairs and Climate Action (BMWK), M-Rock (grant number 01IW21002) funded by the German Federal Ministry for Economic Affairs and Climate Action (BMWK) and the Ministry of Education and Research (BMBF), and ActGPT (grant number 01IW25002) funded by the Federal Ministry of Research, Technology and Space (BMFTR) and is supported with funds from the federal state of Bremen for setting up the Underactuated Robotics Lab (265/004-08-02-02-30365/2024-102966/2024-740847/2024).

\bibliography{references}

\end{document}

%% file: acros.tex
\usepackage[acronym]{glossaries}
\glsdisablehyper

\newacronym{mpc}{MPC}{Model Predictive Control}
\newacronym[plural=QP,firstplural=Quadratic Programs]{qp}{QP}{Quadratic Program}
\newacronym{srbd}{SRBD}{single rigid-body dynamics}
\newacronym{gs}{GS}{Gait Sequencer}
\newacronym{slc}{SLC}{Swing Leg Controller}
\newacronym{wbc}{WBC}{Whole-Body Control}
\newacronym{sfpw}{SFPW}{Solve Frequency per Watt}
\newacronym{cpu}{CPU}{Central Processing Unit}
\newacronym{hw}{HW}{hardware}
\newacronym{com}{COM}{Center of Mass}
\newacronym{ipm}{IPM}{Interior-point method}
\newacronym{asm}{ASM}{Active-set method}
\newacronym{admm}{ADMM}{Alternating direction method of multipliers}
\newacronym{alm}{ALM}{Augmented Lagrangian Method}
\newacronym{csc}{CSC}{Compressed Sparse Column}
\newacronym{pmm}{PMM}{Proximal Method of Multipliers}

\newacronym{fpga}{FPGA}{Field Programmable Gate Arraz}
\newacronym{ags}{AGS}{adaptive Gait Sequencer}
\newacronym{sgs}{SGS}{simple Gait Sequencer}

\newacronym{hpipm}{HPIPM}{High Performance Interior Point Method}
\newacronym{osqp}{OSQP}{Operator Splitting Quadratic Programming Solver}
\newacronym{qpoases}{qpOASES}{Quadratic Programming Online Active Set Solver}
\newacronym{daqp}{DAQP}{Dual Active Set Solver for Quadratic Programming}
\glsunset{hpipm}
\glsunset{osqp}
\glsunset{daqp}
\glsunset{qpoases}


%% file: sections/intro.tex
\begin{figure}
    \centering
    \includegraphics[width=0.98\linewidth]{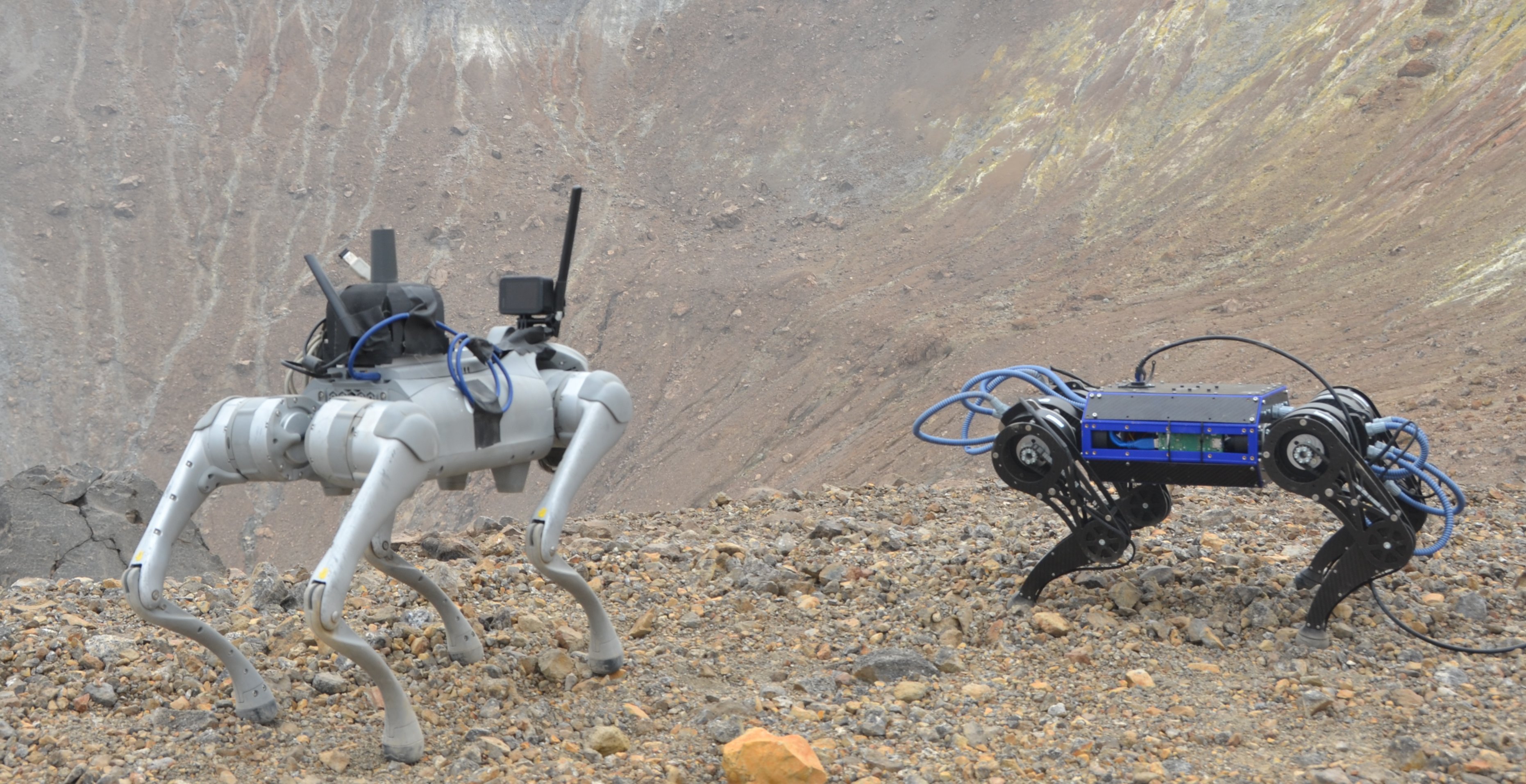}
    \caption{The control framework was successfully tested on both a Unitree Go2 robot and the custom build quadrupedal robot B12 during a field test to Vulcano Island in Italy.}
    \label{fig:quad_vulcano}
\end{figure}

The exploration of celestial bodies like Moon or Mars is one of the most ambitious scientific and technological goals of out time. 
For decades, wheeled rovers have been the state-of-the art for robotic planetary exploration, demonstrating remarkable longevity and success.
However, their mobility is fundamentally limited to relatively flat, obstacle-free terrains, leaving vast regions, such as steep craters, rocky slopes, and loose regolith areas, inaccessible.
Legged robots, particularly quadrupeds, offer a paradigm shift in planetary mobility. 
Inspired by the agility and adaptability of animals, these systems can traverse complex, unstructured environments by dynamically selecting footholds and adjusting their gait and posture. 

Despite their potential, the deployment of legged robots in extraterrestrial environments presents major challenges. 
The control paradigms that enable impressive dynamic motions in laboratory settings often rely on precise knowledge of the robot's dynamic model (e.g., mass, inertia) and its interaction with a known environment. 
In planetary exploration, these parameters are highly uncertain: the robot's mass and center of mass (COM) change as samples are collected and instruments are deployed, and the terrain properties, such as soil composition, slope, and friction, are entirely unknown a priori. 
A control strategy designed for a nominal model will, at best, be suboptimal and, at worst, lead to catastrophic failure in these conditions. 
Consequently, there is a critical need for robust and adaptive control frameworks that can compensate for these uncertainties.

For the field of legged robotics, the state of the art can be categorized into model-based control, learning-based approaches, and recent demonstrations in space analog environments. Yet, what is largely missing is a unifying software and control architecture that can incorporate these advances in a modular, adaptive fashion.
Much of the recent progress in dynamic legged locomotion is built upon hierarchical, model-based controllers. A common architecture is pioneered by work on the Cheetah robots \cite{diCarlo2018Dynamic, kim2019Highly} from MIT.
To overcome the limitations of purely analytical models, data-driven methods have gained prominence. 
Deep Reinforcement Learning (DRL) has been used to train robust locomotion policies in simulation that transfer to real hardware, as demonstrated by systems like ANYmal \cite{ethZurich2023ANYmal}. 
These methods can exhibit remarkable robustness to disturbances and terrain variations. 
Alternatively, bio-inspired approaches directly extract motion patterns from animals \cite{zhang2018ModeAdaptive} to create natural and efficient gaits. 
However, pure learning-based methods can be sample inefficient, lack explicit safety guarantees, and are difficult to verify for safety-critical space applications. 
Our framework leverages the benefits of bio-inspired gait adaptation while retaining the transparency and reliability of model-based control and the online model adaptation compensates for the disadvantages of a faulty model.

The application of legged robots to space exploration is still in its early stages but is an area of intense research and development. 
The German Aerospace Center (DLR) has developed sophisticated quadruped and bipedal robots for space servicing and exploration. NASA's Jet Propulsion Laboratory (JPL) has tested quadruped robots for inspection and exploration tasks. The ESA-supported project ``SpaceBok'' was explicitly designed for dynamic locomotion in low gravity environments. 
Most recently, ETH Zurich's ANYmal has been tested in lunar analog environments \cite{ethZurich2023ANYmal}, demonstrating exceptional mobility. However, these efforts largely demonstrate locomotion capabilities in isolation. 
In contrast, our work emphasizes online adaptation to both terrain and model uncertainty within a unified, modular framework. 
This integrative perspective is crucial for enabling resilient and autonomous operation over long-duration planetary missions.

This work presents such a modular and hierarchical control framework, explicitly designed for runtime adaptability in planetary robotics. The key novelty of our approach is not a single algorithm, but the unification of multiple components, bio-inspired adaptive gait sequencing, real-time MPC, WBC, and online model adaptation, within a software architecture that enables runtime reconfiguration and robust mobility in uncertain environments. These methods excel due to their physical interpretability and performance guarantees, but are inherently limited by the accuracy of the underlying model. Uncertainty in parameters such as mass or center of mass, caused by payload, can significantly degrade performance. While previous work has explored payload identification \cite{tournois2017Online}, these approaches are typically treated as standalone modules and not deeply integrated into a runtime control framework capable of handling significant and persistent uncertainty. We also evaluate the computing requirements of the framework.



%% file: sections/state_estimation.tex
We use the invariant Extended Kalman Filter (InEKF) \cite{inekf2020} to estimate the floating base state of the robot, including the orientation, velocity and position in the world frame. While many approaches rely on a standard EKF for this estimation, we adopt the InEKF because its Lie group formulation naturally respects the underlying geometry of the problem. This provides more consistent and stable error propagation compared to conventional EKF formulations, particularly in scenarios involving non-linear dynamics and contact switching.

Accurate contact detection is critical for floating base estimation, particularly in planetary environments where reliable state feedback underpins mobility and control. We primarily use force or pressure sensors to obtain contact events as they provide direct measurements during nominal operation. However, these signals can drift or become unreliable under environmental effects such as regolith, dust, thermal variation, terrain compliance, or foot slippage.

To ensure robustness, we adopt a hierarchical strategy: pressure sensors are prioritized when their signals are reliable, but when they degrade or not existent, we switch to a momentum observer-based method \cite{bledt2018contact}. The momentum observer exploits the rigid-body dynamics of the robot to estimate the net external forces acting on the system from the joint torques, the kinematics and the dynamics. By monitoring the residual between expected and observed dynamics, it can infer contact events without requiring direct force measurements. This provides a clean, proprioceptively derived contact signal with low delay and resilience against environmental disturbances.

In summary, the contact-aided InEKF provides floating-base estimation by fusing IMU propagation with kinematic constraints, while the momentum observer delivers a proprioceptive signal for contact detection. By combining both modalities, the estimator can exploit direct contact sensing when signals are reliable but can return to momentum-based proprioception when sensors degrade due to regolith, terrain compliance, or thermal effects. This adaptive switching capability ensures consistent state estimation in diverse planetary environments.